\documentclass[letterpaper]{article} 
\usepackage{aaai24}  
\usepackage{times}  
\usepackage{helvet}  
\usepackage{courier}  
\usepackage[hyphens]{url}  
\usepackage{graphicx} 
\usepackage{natbib}  
\usepackage{caption} 
\usepackage{algorithm}
\usepackage{algorithmic}
\usepackage{amsmath}
\usepackage{amsthm}

\newtheorem{theorem}{Theorem}
\usepackage{colortbl}
\usepackage[dvipsnames]{xcolor}

\usepackage{amsmath,amsfonts,bm}

\def\eqref#1{equation~\ref{#1}}

\def\1{\bm{1}}

\def\mP{{\bm{P}}}

\def\mU{{\bm{U}}}
\def\mV{{\bm{V}}}

\DeclareMathAlphabet{\mathsfit}{\encodingdefault}{\sfdefault}{m}{sl}
\SetMathAlphabet{\mathsfit}{bold}{\encodingdefault}{\sfdefault}{bx}{n}

\newcommand{\jiaxing}[1]{{\color{black}#1}}

\usepackage{newfloat}
\usepackage{listings}
\DeclareCaptionStyle{ruled}{labelfont=normalfont,labelsep=colon,strut=off} 
\floatstyle{ruled}
\newfloat{listing}{tb}{lst}{}
\floatname{listing}{Listing}
\title{Union Subgraph Neural Networks}
\author{
    Jiaxing Xu\equalcontrib, Aihu Zhang\equalcontrib, Qingtian Bian, Vijay Prakash Dwivedi, Yiping Ke
}
\affiliations{
    \textsuperscript{\rm }Nanyang Technological University, Singapore\\

    \{jiaxing003,zhan0547,bian0027,vijaypra001\}@e.ntu.edu.sg, ypke@ntu.edu.sg
}

\usepackage{bibentry}

\begin{document}

\maketitle

\begin{abstract}
Graph Neural Networks (GNNs) are widely used for graph representation learning in many application domains. The expressiveness of vanilla GNNs is upper-bounded by 1-dimensional Weisfeiler-Leman (1-WL) test as they operate on rooted subtrees through iterative message passing.
In this paper, we empower GNNs by injecting neighbor-connectivity information extracted from a new type of substructure. We first investigate different kinds of connectivities existing in a local neighborhood and identify a substructure called union subgraph, which is able to capture the complete picture of the 1-hop neighborhood of an edge. We then design a shortest-path-based substructure descriptor that possesses three nice properties and can effectively encode the high-order connectivities in union subgraphs. By infusing the encoded neighbor connectivities, we propose a novel model, namely Union Subgraph Neural Network (UnionSNN), which is proven to be strictly more powerful than 1-WL in distinguishing non-isomorphic graphs. Additionally, the local encoding from union subgraphs can also be injected into arbitrary message-passing neural networks (MPNNs) and Transformer-based models as a plugin. Extensive experiments on \jiaxing{18} benchmarks of both graph-level and node-level tasks demonstrate that UnionSNN outperforms state-of-the-art baseline models, with competitive computational efficiency. The injection of our local encoding to existing models is able to boost the performance by up to 11.09\%. \jiaxing{Our code is available at \url{https://github.com/AngusMonroe/UnionSNN}.}
\end{abstract}

\section{Introduction}
With the ubiquity of graph-structured data emerging from various modern applications, Graph Neural Networks (GNNs) have gained increasing attention from both researchers and practitioners. GNNs have been applied to many application domains, including quantum chemistry \cite{duvenaud2015convolutional,dai2016discriminative,masters2022gps++}, social science \cite{ying2018hierarchical,fan2019graph,shiao2022link}, transportation \cite{peng2020spatial,derrow2021eta} and neuroscience \cite{ahmedt2021graph, wang2021scgnn}, and have attained promising results on graph classification \cite{ying2018hierarchical,masters2022gps++}, node classification \cite{shi2020masked} and link prediction \cite{zhang2018link,suresh2023expressive} tasks. 

Most GNNs are limited in terms of their expressive power. Xu et al., \cite{xu2018powerful} show that GNNs are at most as powerful as 1-dimentional Weisfeiler-Leman (1-WL) test \cite{weisfeiler1968reduction} in distinguishing non-isomorphic graph structures. This is because a vanilla GNN essentially operates on a subtree rooted at each node in its message passing, \textit{i.e.}, it treats every neighbor of the node equally in its message aggregation. In this regard, it overlooks any discrepancy that may exist in the connectivities between neighbors. To address this limitation, efforts have been devoted to incorporating local substructure information to GNNs. Several studies attempt to encode such local information through induced subgraphs \cite{zhao2021stars}, overlap subgraphs \cite{wijesinghe2021new} and spatial encoding \cite{bouritsas2022improving}, to enhance GNNs' expressiveness. But the local structures they choose are not able to capture the complete picture of the 1-hop neighborhood of an edge. Some others incorporate shortest path information to edges in message passing via distance encoding \cite{li2020distance}, adaptive breath/depth functions \cite{liu2019geniepath} and affinity matrix \cite{wan2021principled} to control the message from neighbors at different distances. However, the descriptor used to encode the substructure may overlook some connectivities between neighbors. Furthermore, some of the above models also suffer from high computational cost due to the incorporation of certain substructures. 

In this paper, we aim to develop a model that overcomes the above drawbacks and yet is able to empower GNNs' expressiveness. (1) We define a new type of substructures named union subgraphs, each capturing the entire closed neighborhood w.r.t. an edge. (2) We design an effective substructure descriptor that encodes high-order connectivities and it is easy to incorporate with arbitrary message-passing neural network (MPNN) or Transformer-based models. (3) We propose a new model, namely Union Subgraph Neural Network (UnionSNN), which is strictly more expressive than the vanilla GNNs (1-WL) in theory and also computationally efficient in practice. Our contributions are summarized as follows:

\begin{itemize}
\item We investigate different types of connectivities existing in the local neighborhood and identify the substructure, named ``union subgraph'', that is able to capture the complete 1-hop neighborhood.  
\item We abstract three desirable properties for a good substructure descriptor and design a shortest-path-based descriptor that possesses all the properties with high-order connectivities encoded. 
\item We propose a new model, UnionSNN, which incorporates the information extracted from union subgraphs into the message passing. We also show how our local encoding can be flexibly injected into any arbitrary MPNNs and Transformer-based models.  We theoretically prove that UnionSNN is more expressive than 1-WL. We also show that UnionSNN is stronger than 3-WL in some cases.
\item We perform extensive experiments on both graph-level and node-level tasks. UnionSNN consistently outperforms baseline models on \jiaxing{18} benchmark datasets, with competitive efficiency. The injection of our local encoding is able to boost the performance of base models by up to 11.09\%, which justifies the effectiveness of our proposed union subgraph and substructure descriptor in capturing local information. 
\end{itemize}

\section{Related Work}
\label{gen_inst}

\subsection{Substructure-Enhanced GNNs}

In recent years, several GNN architectures have been designed to enhance their expressiveness by encoding local substructures. GraphSNN \cite{wijesinghe2021new} brings the information of overlap subgraphs into the message passing scheme as a structural coefficient. However, the overlap subgraph and the substructure descriptor used by GraphSNN are not powerful enough to distinguish all non-isomorphic substructures in the 1-hop neighborhood. Zhao et al. \cite{zhao2021stars} encode the induced subgraph for each node and inject it into node representations. \cite{bouritsas2022improving} introduces structural biases in the aggregation function to break the symmetry in message passing. For these two methods, the neighborhood under consideration should be pre-defined, and the subgraph matching is extremely expensive ($O(n^k)$ for $k$-tuple substructure) when the substructure gets large. 
Similarly, a line of research \cite{bodnar2021weisfeiler,thiede2021autobahn,horn2021topological} develops new WL aggregation schemes to take into account substructures like cycles or cliques. 
Despite these enhancements, performing cycle counting is very time-consuming. Other Transformer-based methods \cite{dwivedi2020generalization,kreuzer2021rethinking,wu2021representing,mialon2021graphit} incorporate local structural information via positional encoding \cite{lim2022sign,zhang2023rethinking}. Graphormer \cite{ying2021graphormer} combines the node degree and the shortest path information for spatial encoding, 
while other works \cite{li2020distance,dwivedi2021graph} employ random walk based encodings that can encode $k$-hop neighborhood information of a node. However, these positional encodings only consider relative distances from the center node and ignore high-order connectivities between the neighbors.

\subsection{Path-Related GNNs}

A significant amount of works have focused on the application of shortest paths and their related techniques to GNNs. 
Li et al. \cite{li2020distance} present a distance encoding module to augment node features and control the receptive field of message passing.
GeniePath \cite{liu2019geniepath} proposes an adaptive breath function to learn the importance of different-sized neighborhoods and an adaptive depth function to extract and filter signals from neighbors within different distances. PathGNN \cite{tang2020towards} imitates how the Bellman-Ford algorithm solves the shortest path problem in generating weights when updating node features. 
SPN \cite{abboud2022shortest} designs a scheme, in which the representation of a node is propagated to each node in its shortest path neighborhood. Some recent works adapt the concept of curvature from differential geometry to reflect the connectivity between nodes and the possible bottleneck effects. CurvGN \cite{ye2019curvature} reflects how easily information flows between two nodes by graph curvature information, and exploits curvature to reweigh different channels of messages. Topping et al. \cite{topping2021understanding} propose Balanced Forman curvature that better reflects the edges having bottleneck effects, and alleviate the over-squashing problem of GNNs by rewiring graphs. SNALS \cite{wan2021principled} utilizes an affinity matrix based on shortest paths to encode the structural information of hyperedges.
Our method is different from these existing methods by introducing a shortest-path-based substructure descriptor for distinguishing non-isomorphic substructures.

\section{Local Substructures to Empower MPNNs}

In this section, we first introduce MPNNs. We then investigate what kind of local substructures are beneficial to improve the expressiveness of MPNNs.

\subsection{Message Passing Neural Networks}

We represent a graph as $G = (V, E, X)$, where $V=\{v_1, ..., v_n\}$ is the set of nodes, $E \in V \times V$ is the set of edges, and $X=\left\{\mathbf{x}_v \mid v \in V\right\} $ is the set of node features. The set of neighbors of node $v$ is denoted by $\mathcal{N}(v)=\{u \in V \mid(v, u) \in E\}$. 
The $l$-th layer of an MPNN \cite{xu2018powerful} can be written as:
\begin{equation}
\scriptsize
\mathbf{h}_v^{(l)}=\operatorname{AGG}^{(l-1)}(\mathbf{h}_v^{(l-1)}, \operatorname{MSG}^{(l-1)}(\{\mathbf{h}_u^{(l-1)}, u \in \mathcal{N}(v)\})),
\label{eq:mpnn}
\end{equation}
\noindent
where $\mathbf{h}_v^{(l)}$ is the representation of node $v$ at the $l$-th layer, $\mathbf{h}_v^{(0)} = \mathbf{x}_v$, $\operatorname{AGG}(\cdot)$ and $\operatorname{MSG}(\cdot)$ denote the aggregation and message functions, respectively.

\subsection{Local Substructures to Improve MPNNs}
\label{sec:structural}

\begin{figure*}[ht]
\begin{center}
\includegraphics[width=.6\linewidth]{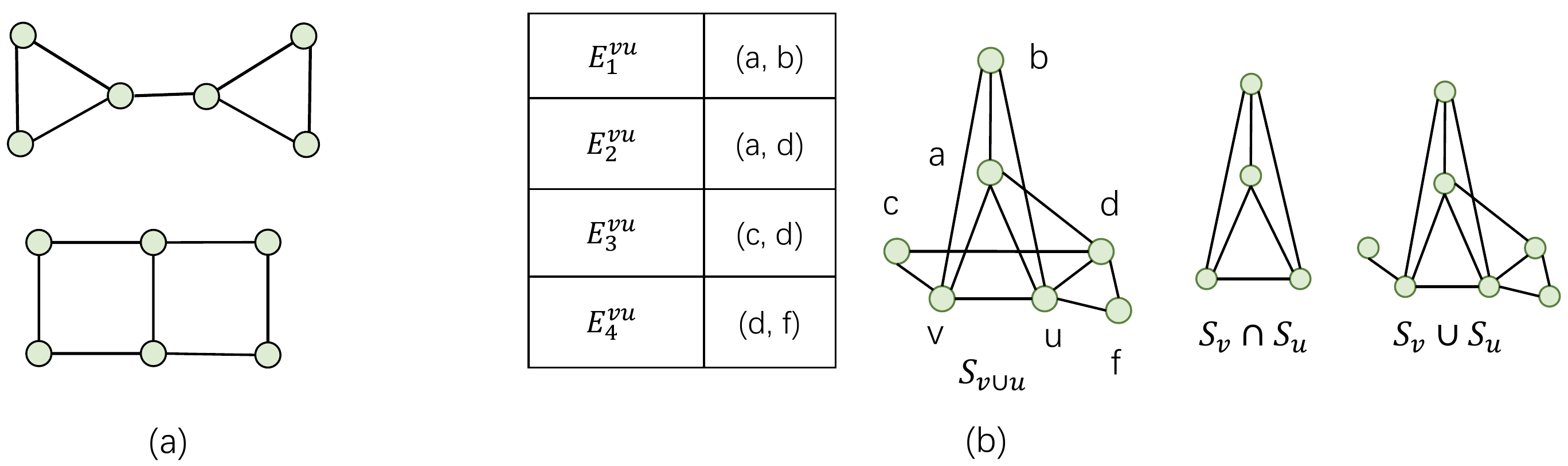}
\end{center}
\caption{(a) A pair of non-isomorphic graphs not distinguishable by 1-WL; (b) An example of various local substructures for two adjacent nodes $v$ and $u$.}
\label{fig:structural_analysis}
\end{figure*}

According to Eq. (\ref{eq:mpnn}), MPNN updates the representation of a node isotropously at each layer and ignores the structural connections between the neighbors of the node. Essentially, the local substructure utilized in the message passing of MPNN is a subtree rooted at the node. Consequently, if two non-isomorphic graphs have the same set of rooted subtrees, they cannot be distinguished by MPNN (and also 1-WL). Such an example is shown in Figure \ref{fig:structural_analysis}(a). A simple fix to this problem is to encode the local structural information about each neighbor, based on which neighbors are treated unequally in the message passing. One natural question arises: \textbf{which substructure shall we choose to characterize the 1-hop local information?}

To answer the above question, we consider two adjacent nodes $v$ and $u$, and discuss different types of edges that may exist in their neighbor sets, $\mathcal{N}(v)$ and $\mathcal{N}(u)$. We define the closed neighbor set of node $v$ as $\tilde{\mathcal{N}}(v) = \mathcal{N}(v) \cup \{v\}$. The induced subgraph of $\tilde{\mathcal{N}}(v)$ is denoted by $S_v$, which defines the closed neighborhood of $v$. The common closed neighbor set of $v$ and $u$ is $\mathcal{N}_{vu} = \tilde{\mathcal{N}}(v) \cap \tilde{\mathcal{N}}(u)$ and the exclusive neighbor set of $v$ w.r.t $u$ is defined as $\mathcal{N}_v^{-u} = \tilde{\mathcal{N}}(v) - \mathcal{N}_{vu}$. As shown in Figure \ref{fig:structural_analysis}(b), there are four types of edges in the closed neighborhood of $\{v, u\}$:

\begin{itemize}
\item \textbf{$E_1^{vu} \in \mathcal{N}_{vu} \times \mathcal{N}_{vu}$}: edges between the common closed neighbors of $v$ and $u$, such as $(a, b)$;
\item \textbf{$E_2^{vu} \in (\mathcal{N}_{vu} \times \mathcal{N}_{v}^{-u}) \cup (\mathcal{N}_{vu} \times \mathcal{N}_{u}^{-v})$}: edges between a common closed neighbor of $v$ and $u$ and an exclusive neighbor of $v$/$u$, such as $(a, d)$;
\item \textbf{$E_3^{vu} \in \mathcal{N}_{v}^{-u} \times \mathcal{N}_{u}^{-v}$}: edges between two exclusive neighbors of $v$ and $u$ from different sides, such as $(c, d)$;
\item \textbf{$E_4^{vu} \in (\mathcal{N}_{v}^{-u} \times \mathcal{N}_{v}^{-u}) \cup (\mathcal{N}_{u}^{-v} \times \mathcal{N}_{u}^{-v})$}: edges between two exclusive neighbors of $v$ or $u$ from the same side, such as $(d, f)$.
\end{itemize}

We now discuss three different local substructures, each capturing a different set of edges.

\noindent
\textbf{Overlap Subgraph\cite{wijesinghe2021new}.} The overlap subgraph of two adjacent nodes $v$ and $u$ is defined as $S_{v \cap u} = S_v \cap S_u$. The overlap subgraph contains only edges in $E_1^{vu}$.

\noindent
\textbf{Union Minus Subgraph.} The union minus subgraph of two adjacent nodes $v$, $u$ is defined as $S_{v \cup u}^- = S_v \cup S_u$. The union minus subgraph consists of edges in $E_1^{vu}$, $E_2^{vu}$ and $E_4^{vu}$.

\noindent
\textbf{Union Subgraph}. The union subgraph of two adjacent nodes $v$ and $u$, denoted as $S_{v \cup u}$, is defined as the induced subgraph of $\tilde{\mathcal{N}}(v) \cup \tilde{\mathcal{N}}(u)$. The union subgraph contains all four types of edges mentioned above.

It is obvious that union subgraph captures the whole picture of the 1-hop neighborhood of two adjacent nodes. 
This subgraph captures all types of connectivities within the neighborhood, providing an ideal local substructure for enhancing the expressive power of MPNNs.
We illustrate how effective different local substructures are in improving MPNNs through an example in Appendix \ref{app:substructures}. 
Note that we restrict the discussion to the 1-hop neighborhood because we aim to develop a model based on the MPNN scheme, in which a single layer of aggregation is performed on the 1-hop neighbors.

\subsection{Union Isomorphism}
\label{sec:union_isomorphism}

We now proceed to define the isomorphic relationship between the neighborhoods of two nodes $i$ and $j$ based on union subgraphs. The definition follows that of overlap isomorphism in \cite{wijesinghe2021new}. 

\noindent
\textbf{Overlap Isomorphism}. $S_i$ and $S_j$ are overlap-isomorphic, denoted as $S_i \simeq_{overlap} S_j$, if there exists a bijective mapping $g$: $\tilde{\mathcal{N}}(i) \rightarrow \tilde{\mathcal{N}}(j)$ such that $g(i) = j$, and for any $v \in \mathcal{N}(i)$ and $g(v) = u$, $S_{i \cap v}$ and $S_{j \cap u}$ are isomorphic (ordinary graph isomorphic).

\noindent
\textbf{Union Isomorphism}. $S_i$ and $S_j$ are union-isomorphic, denoted as $S_i \simeq_{union} S_j$, if there exists a bijective mapping $g$: $\tilde{\mathcal{N}}(i) \rightarrow \tilde{\mathcal{N}}(j)$ such that $g(i) = j$, and for any $v \in \mathcal{N}(i)$ and $g(v) = u$, $S_{i \cup v}$ and $S_{j \cup u}$ are isomorphic (ordinary graph isomorphic).

\begin{figure}[h]
\begin{center}
\includegraphics[width=.9\linewidth]{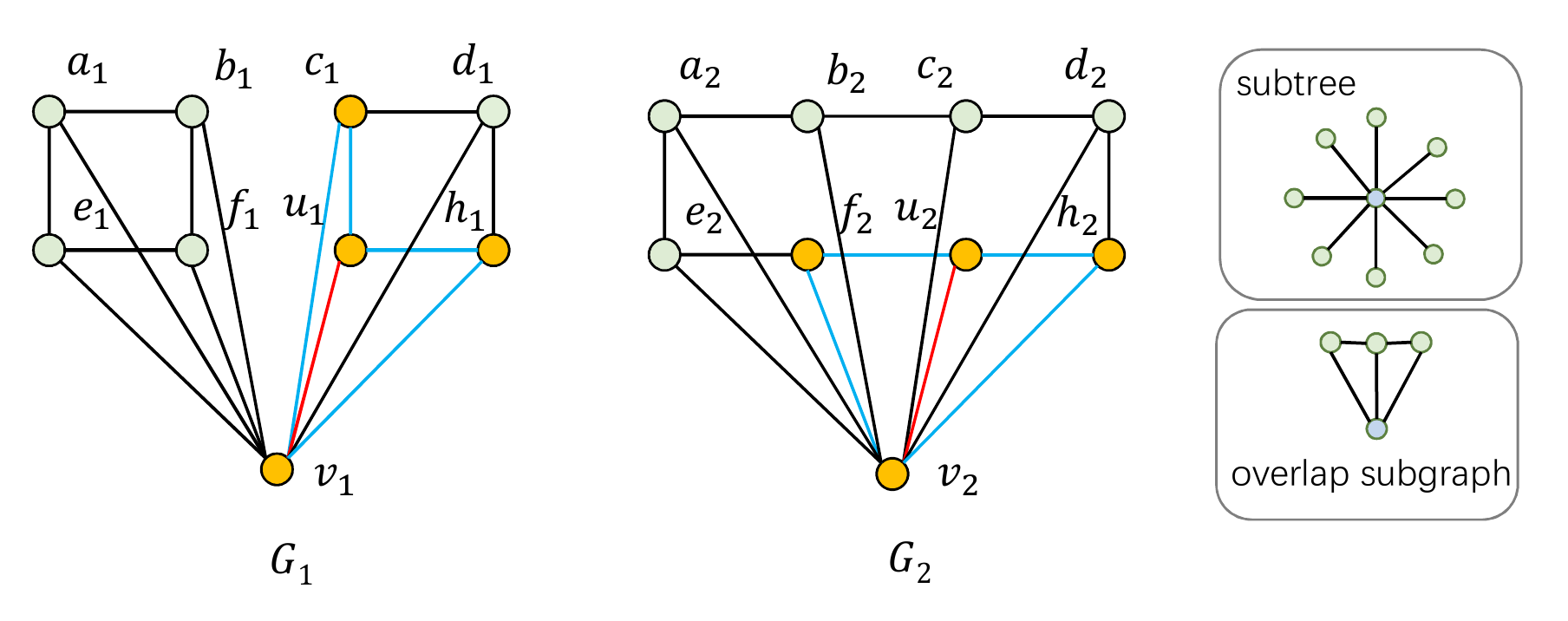}
\end{center}
\caption{An example that the bijective mapping between the nodes in the subgraphs is not the same as $g$.}
\label{fig:isoexample}
\end{figure}

As shown in Figure \ref{fig:isoexample}, we have two subgraphs $G_1$ and $G_2$. These two subgraphs are overlap-isomorphic. The bijective mapping $g$ is: $g(k_1) = g(k_2)$, $\forall k \in \{v, a, b, c, d, e, f, u, h\}$. Take a pair of corresponding overlap subgraphs as an example: $S_{v_1 \cap u_1}$ and $S_{v_2 \cap u_2}$. They are based on two red edges $(v_1, u_1)$, and $(v_2, u_2)$ and the rest of edges in the overlap subgraphs are colored in blue. It is easy to see that $S_{v_1 \cap u_1}$ and $S_{v_2 \cap u_2}$ are isomorphic (ordinary one). In this ordinary graph isomorphism, its bijective mapping between the nodes is not required to be the same as $g$. It could be the same or different, as long as the ordinary graph isomorphism holds between this pair of overlap subgraphs. In this example, any pair of corresponding overlap subgraphs defined under $g$ are isomorphic (ordinary one). In fact, all overlap subgraphs have the same structure in this example. In this sense, the concept of “overlap isomorphism” does not look at the neighborhood based on a single edge, but captures the neighborhoods of all edges and thus the “overlap” connectivities of the two subgraphs $G_1$ and $G_2$. As for the union subgraph concept we define, the union subgraph of nodes $v_1$ and $u_1$ ($S_{v_1 \cup u_1}$) is the same as $G_1$, and the union subgraph of nodes $v_2$ and $u_2$ ($S_{v_2 \cup u_2}$) is the same as $G_2$. Therefore, $S_{v_1 \cup u_1}$ and $S_{v_2 \cup u_2}$ are not union-isomorphic. This means that union isomorphism is able to distinguish the local structures centered at $v_1$ and $v_2$ but overlap isomorphism cannot.

\begin{theorem}
\label{thm1}
If $S_i \simeq_{union} S_j$, then $S_i \simeq_{overlap} S_j$; but not vice versa.
\end{theorem}

Theorem \ref{thm1} states that union-isomorphism is stronger than overlap-isomorphism. The proofs of all theorems are provided in Appendix \ref{app:proofs}.

\section{UnionSNN}


\subsection{Design of Substructure Descriptor Function}
\label{sec:descriptor}

Let $\mathcal{U} = \{S_{v \cup u} | (v, u) \in E\}$ be the set of union subgraphs in $G$. In order to fuse the information of union subgraphs in message passing, we need to define a function $f(\cdot)$ to describe the structural information of each $S_{v \cup u} \in \mathcal{U}$. Ideally, given two union subgraphs centered at node $v$, $S_{v \cup u} = (V_{v \cup u}, E_{v \cup u})$ and $S_{v \cup u^\prime} = (V_{v \cup u^\prime}, E_{v \cup u^\prime})$, we want $f(S_{v \cup u}) = f(S_{v \cup u^\prime})$ iff $S_{v \cup u}$ and $S_{v \cup u^\prime}$ are isomorphic. We abstract the following properties of a good substructure descriptor function $f(\cdot)$:

\begin{figure*}[h]
\begin{center}
\includegraphics[width=.70\linewidth]{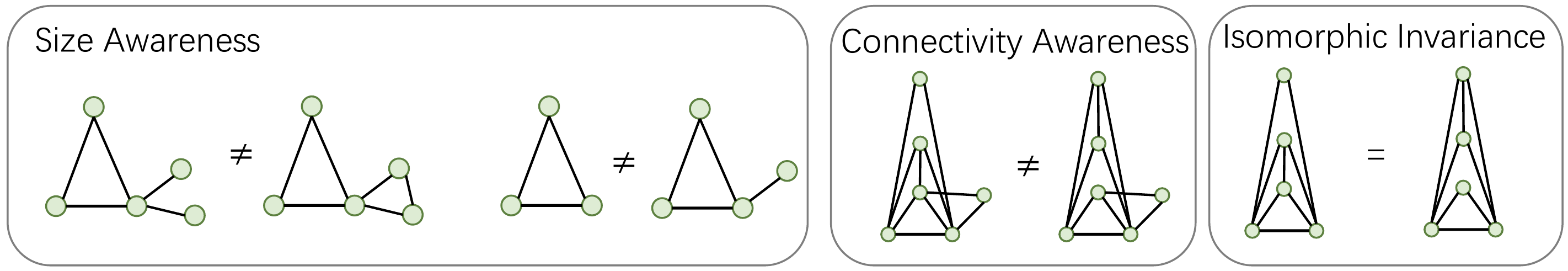}
\end{center}
\caption{Three properties that a good substructure descriptor function $f(\cdot)$ should exhibit.}
\label{fig:properties}
\end{figure*}

\begin{itemize}
\item \textbf{Size Awareness}. $f(S_{v \cup u}) \neq f(S_{v \cup u^\prime})$ if $|V_{v \cup u}| \neq |V_{v \cup u^\prime}|$ or $|E_{v \cup u}| \neq |E_{v \cup u^\prime}|$; 
\item \textbf{Connectivity Awareness}. $f(S_{v \cup u}) \neq f(S_{v \cup u^\prime})$ if $|V_{v \cup u}| = |V_{v \cup u^\prime}|$ and $|E_{v \cup u}| = |E_{v \cup u^\prime}|$ but $S_{v \cup u}$ and $S_{v \cup u^\prime}$ are not isomorphic;
\item \textbf{Isomorphic Invariance}. $f(S_{v \cup u}) = f(S_{v \cup u^\prime})$ if $S_{v \cup u}$ and $S_{v \cup u^\prime}$ are isomorphic.
\end{itemize}

Figure \ref{fig:properties} illustrates the properties. Herein, we design $f(\cdot)$ as a function that transforms $S_{v \cup u}$ to a path matrix $\bm{P}^{vu} \in \mathbb{R}^{|V_{v \cup u}| \times |V_{v \cup u}|}$ such that each entry:
\begin{equation}
\small
\bm{P}_{ij}^{vu} = \operatorname{PathLen}(i, j, S_{v \cup u}), i, j \in V_{v\cup u},
\label{eq:path_matrix}
\end{equation}
\noindent
where $\operatorname{PathLen}(\cdot)$ denotes the length of the shortest path between $i$ and $j$ in $S_{v \cup u}$. We choose the path matrix over the adjacency matrix or the Laplacian matrix as it explicitly encodes high-order connectivities between the neighbors. In addition, with a fixed order of nodes, we can get a unique $\bm{P}^{vu}$ for a given $S_{v \cup u}$, and vice versa. We formulate it in Theorem \ref{thm2}.  

\begin{theorem}
\label{thm2}
With a fixed order of nodes in the path matrix, we can obtain a unique path matrix $\bm{P}^{vu}$ for a given union subgraph $S_{v \cup u}$, and vice versa.
\end{theorem}

It is obvious that our proposed $f(\cdot)$ satisfies the above-mentioned three properties, with a node permutation applied in the isomorphic case. 

\noindent
\textbf{Discussion on Other Substructure Descriptor Functions}. In the literature, some other functions have also been proposed to describe graph substructures. (1) Edge Betweenness \cite{brandes2001faster} is defined by the number of shortest paths between any pair of nodes in a (sub)graph $G$ that pass through an edge. When applying the edge betweenness to $(v, u)$ in $S_{v \cup u}$, the metric would remain the same on two different union subgraphs, one with an edge in $E_4^{vu}$ and one without. This shows that edge betweenness does not satisfy Size Awareness; (2) Wijesinghe and Wang \cite{wijesinghe2021new} puts forward a substructure descriptor as a function of the number of nodes and edges. This descriptor fails to distinguish non-isomorphic subgraphs with the same size, and thus does not satisfy Connectivity Awareness; (3) Discrete Graph Curvature, e.g., Olliveier Ricci curvature \cite{ollivier2009ricci,lin2011ricci}, has been introduced to MPNNs in recent years \cite{ye2019curvature}. Ricci curvature first computes for each node a probability vector of length $|V|$ that characterizes a uniform propagation distribution in the neighborhood. It then defines the curvature of two adjacent nodes as the Wasserstein distance of their corresponding probability vectors. Similar to edge betweenness,  curvature doesn't take into account the edges in $E_4^{vu}$ in its computation and thus does not satisfy Size Awareness either.
We detail the definitions of these substructure descriptor functions in Appendix \ref{app:sub_des_func}.

\subsection{Network Design}
\label{sec:network}

For the path matrix of an edge $(v, u)$ to be used in message passing, we need to further encode it as a scalar. We perform Singular Value Decomposition (SVD) \cite{horn2012matrix} on the path matrix and extract the singular values: $\bm{P}=\bm{U} \boldsymbol{\Sigma} \bm{V}^*$.
The sum of the singular values of $\bm{P}^{vu}$, denoted as $a^{vu} = \operatorname{sum}(\boldsymbol{\Sigma}^{vu})$ , is used as the local structural coefficient of the edge $(v, u) \in E$. Note that since the local structure never changes in message passing, we can compute the structural coefficients in preprocessing before the training starts. A nice property of this structural coefficient is that, it is \textbf{permutation invariant} thanks to the use of SVD and the sum operator. With arbitrary order of nodes, the computed $a^{vu}$ remains the same, which removes the condition required by Theorem \ref{thm2}.

\noindent
\textbf{UnionSNN.} We now present our model, namely Union Subgraph Neural Network (UnionSNN), which utilizes union-subgraph-based structural coefficients to incorporate local substructures in message passing. For each vertex $v \in V$, the node representation at the $l$-th layer is generated by:
\begin{equation}
\begin{split}
\small
\bm{h}_v^{(l)}=\operatorname{MLP_1}^{(l-1)}((1+\epsilon^{(l-1)}) \bm{h}_v^{(l-1)} + \\ \sum_{u \in \mathcal{N}(v)} \operatorname{Trans}^{(l-1)}(\tilde{a}^{vu})\bm{h}_u^{(l-1)}),
\end{split}
\label{eq:UnionSNN}
\end{equation}
\noindent
where $\epsilon^{(l-1)}$ is a learnable scalar parameter and $\tilde{a}^{vu} = \frac{a^{vu}}{\sum_{u \in \mathcal{N}(v)}a^{vu}}$. $\operatorname{MLP_1}(\cdot)$ denotes a multilayer perceptron (MLP) with a non-linear function ReLU. To transform the weight $\tilde{a}^{vu}$ to align with the multi-channel representation $\bm{h}_u^{(l-1)}$, we follow \cite{ye2019curvature} and apply a transformation function $\operatorname{Trans}(\cdot)$ for better expressiveness and easier training:
\begin{equation}
\small
\operatorname{Trans}(a) = \operatorname{softmax}(\operatorname{MLP_2}(a)),
\label{eq:mlp2}
\end{equation}
\noindent
where $\operatorname{MLP_2}$ denotes an MLP with ReLU and a channel-wise softmax function $\operatorname{softmax}(\cdot)$ normalizes the outputs of MLP separately on each channel.
For better understanding, we provide the pseudo-code of UnionSNN in Appendix \ref{app:code}.

\noindent
\textbf{As a Plugin to Empower Other Models.} In addition to a standalone UnionSNN network, our union-subgraph-based structural coefficients could also be incorporated into other GNNs in a flexible and yet effective manner. For arbitrary MPNNs as in Eq. (\ref{eq:mpnn}), we can plugin our structural coefficients via an element-wise multiplication:
\begin{equation}
\begin{split}
\small
\mathbf{h}_v^{(l)}=\operatorname{AGG}^{(l-1)}(\mathbf{h}_v^{(l-1)}, \operatorname{MSG}^{(l-1)}( \\ \{\operatorname{Trans}^{(l-1)}(\tilde{a}^{vu})\mathbf{h}_u^{(l-1)}, u \in \mathcal{N}(v)\})).
\end{split}
\label{eq:unionmpnn}
\end{equation}

For transformer-based models, inspired by the spatial encoding in Graphormer \cite{ying2021graphormer}, we inject our structural coefficients into the attention matrix as a bias term:

\begin{equation}
\small
A_{v u}=\frac{\left(h_v W_Q\right)\left(h_u W_K\right)^T}{\sqrt{d}}+\operatorname{Trans}(\tilde{a}^{vu}),
\label{eq:uniontransformer}
\end{equation}

\noindent
where the definition of $\operatorname{Trans(\cdot)}$ is the same as Eq. (\ref{eq:mlp2}) and shared across all layers, $h_v, h_u \in \mathbb{R}^{1 \times d}$ are the node representations of $v$ and $u$, $W_Q, W_K \in \mathbb{R}^{d \times d}$ are the parameter matrices, and $d$ is the hidden dimension of $h_v$ and $h_u$.

\begin{figure}[h]
\centering
\includegraphics[width=.35\linewidth]{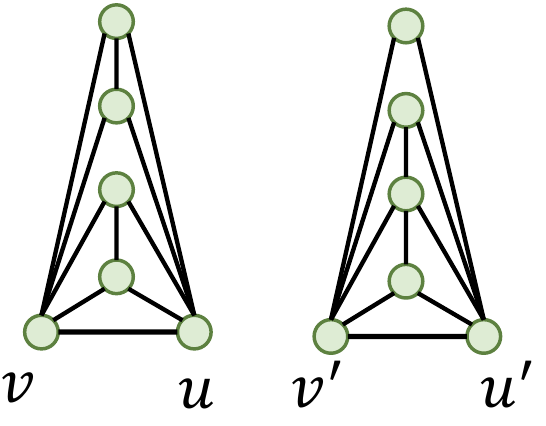}
\caption{Example of two non-isomorphic subgraphs with the same structural coefficient in GraphSNN}
\label{fig:overlap_case}
\end{figure}

\noindent
\textbf{Design Comparisons with GraphSNN.} Our UnionSNN is similar to GraphSNN in the sense that both improve the expressiveness of MPNNs (and 1-WL) by injecting the information of local substructures. However, UnionSNN is superior to GraphSNN in the following aspects. (1) Union subgraphs in UnionSNN are stronger than overlap subgraphs in GraphSNN, as ensured by Theorem \ref{thm1}. (2) The shortest-path-based substructure descriptor designed in UnionSNN is more powerful than that in GraphSNN: the latter fails to possess the property of Connectivity Awareness (as elaborated in Section \ref{sec:descriptor}). An example of two non-isomorphic subgraphs $S_{v \cap u}$ and $S_{v^\prime \cap u^\prime}$ is shown in Figure \ref{fig:overlap_case}. They have the same structural coefficients in GraphSNN. (3) The aggregation function in UnionSNN works on adjacent nodes in the input graph, while that in GraphSNN utilizes the structural coefficients on all pairs of nodes (regardless of their adjacency). Consequently, GraphSNN requires to pad the adjacency matrix and feature matrix of each graph to the maximum graph size, which significantly increases the computational complexity. The advantages of UnionSNN over GraphSNN are also evidenced by experimental results in Section \ref{sec:eff_exp}.

\noindent
\jiaxing{\textbf{Time and Space Complexities.} Given that the path matrices have been precomputed, our time and space complexities for model training are in line with those of GIN and GraphSNN. Denote $m$ as the number of edges in a graph, $f$ and $d$ as the dimensions of input and output feature vectors, and $k$ as the number of layers. Time and space complexities of UnionSNN are $\mathcal{O}(kmfd)$ and $\mathcal{O}(m)$, respectively.}

\subsection{Expressive Power of UnionSNN}

We formalize the following theorem to show that UnionSNN is more powerful than 1-WL test in terms of expressive power.

\begin{theorem}
\label{thm3}
UnionSNN is more expressive than 1-WL in testing non-isomorphic graphs.
\end{theorem}

The stronger expressiveness of UnionSNN over 1-WL is credited to its use of union subgraphs, with an effective encoding of local neighborhood connectivities via the shortest-path-based design of structural coefficients. 

\begin{table*}[h]
\centering
\scalebox{0.8}{
\begin{tabular}{lccccc}
\hline
\textbf{Dataset} &
  \multicolumn{1}{l}{\textbf{Graph \#}} &
  \multicolumn{1}{c}{\textbf{Class \#}} &
  \multicolumn{1}{c}{\textbf{Avg Node \#}} &
  \multicolumn{1}{c}{\textbf{Avg Edge \#}} &
  \multicolumn{1}{c}{\textbf{Metric}} \\\hline
MUTAG        & 188  & 2 & 17.93  & 19.79  & Accuracy \\
PROTEINS     & 1113 & 2 & 39.06  & 72.82  & Accuracy \\
ENZYMES      & 600  & 6 & 32.63  & 62.14  & Accuracy \\  
DD         & 1178 & 2 & 284.32 & 715.66 & Accuracy \\
FRANKENSTEIN & 4337 & 2 & 16.90  & 17.88 & Accuracy \\
Tox21        & 8169 & 2 & 18.09  & 18.50  & Accuracy \\
NCI1         & 4110 & 2 & 29.87  & 32.30 & Accuracy \\
NCI109       & 4127 & 2 & 29.68  & 32.13 & Accuracy \\
OGBG-MOLHIV & 41127 & 2 & 25.50  & 27.50 & ROC-AUC \\ 
OGBG-MOLBBBP   & 2039 & 2 & 24.06 & 25.95 & ROC-AUC \\
ZINC10k   & 12000 & - & 23.16 & 49.83 & MAE \\
ZINC-full   & 249456 & - & 23.16 & 49.83 & MAE \\
4CYCLES  & 20000 & 2 & 36.00 & 61.71 & Accuracy \\
6CYCLES  & 20000 & 2 & 56.00 & 87.84 & Accuracy \\\hline
\end{tabular}
}
\caption{Statistics of Graph Classification/Regression Datasets.}
\label{tab:graph_dataset}
\end{table*}

\begin{table*}[h]
\centering
\scalebox{0.8}{
\begin{tabular}{l|ccccc}
\hline
         & \textbf{Node \#} & \textbf{Edge \#} & \textbf{Class \#} & \textbf{Feature \#} & \textbf{Training \#} \\ \hline
Cora     & 2708   & 5429   & 7       & 1433      & 140        \\
Citeseer & 3327   & 4732   & 6       & 3703      & 120        \\
PubMed   & 19717  & 44338  & 3       & 500       & 60         \\ 
Computer & 13381  & 259159 & 10      & 7667       & 1338     \\
Photo    & 7487   & 126530 & 8       & 745       & 749       \\\hline
\end{tabular}
}
\caption{Statistics of Node Classification Datasets.}
\label{tab:node_dataset}
\end{table*}

\begin{figure}[h]
\begin{center}
\includegraphics[width=.75\linewidth]{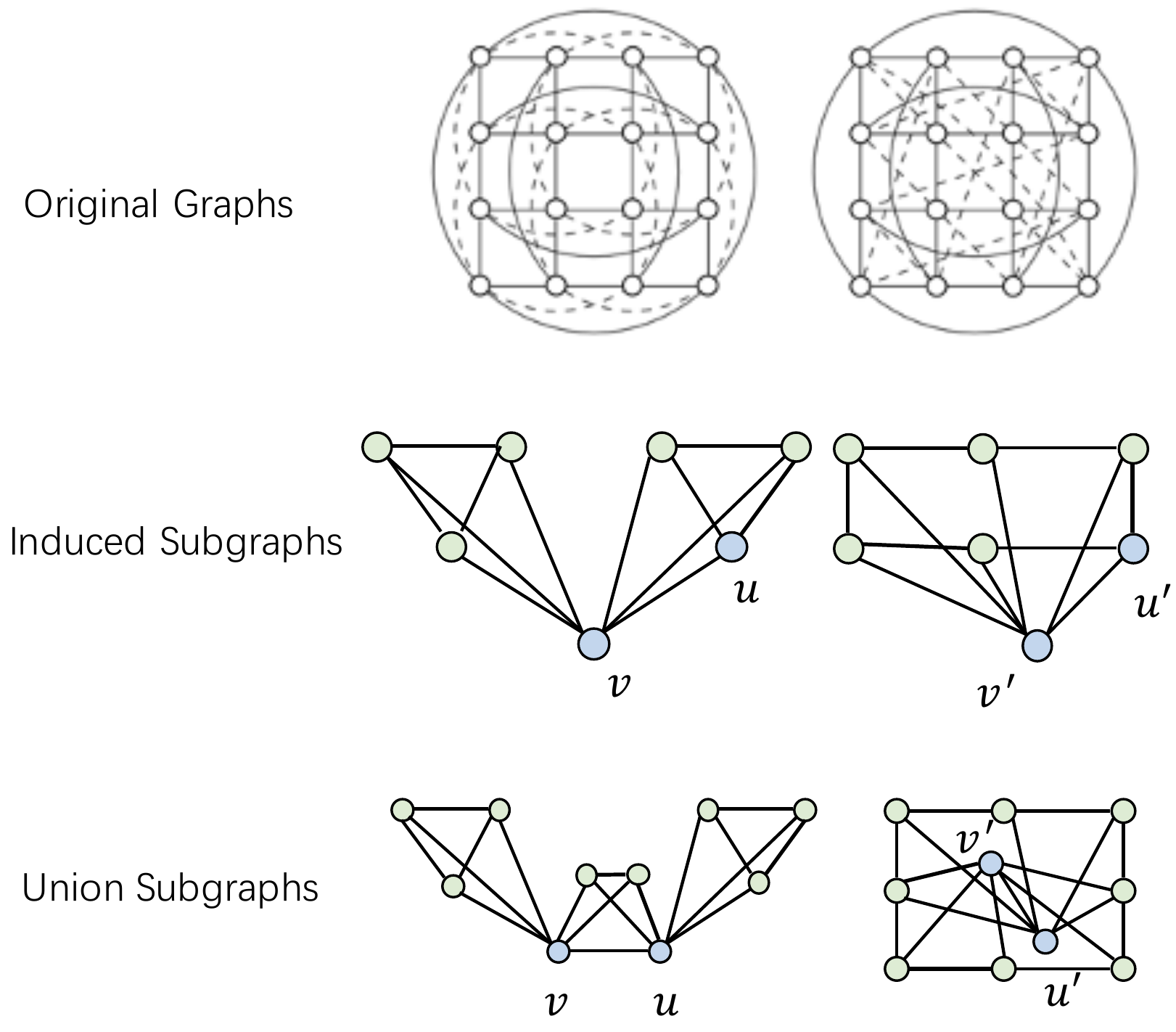}
\end{center}
\caption{These two graphs can be distinguished by UnionSNN but not by 3-WL.}
\label{fig:non-iso_graphs}
\end{figure}

\noindent
\textbf{The Connection with Higher-Order WL Tests.} As discussed in \cite{papp2022theoretical}, high-order WL tests are concepts of global comparisons over two graphs. Therefore, it is arguable if the WL hierarchy is a suitable tool to measure the expressiveness of GNN extensions as the latter focus on locality. Nonetheless, there exist some graph structures on which the 3-WL test is not stronger than UnionSNN, i.e., some graphs can be distinguished by UnionSNN but not by 3-WL. As an example, UnionSNN can distinguish the 4x4 Rook’s graph and the Shrikhande graph (as shown in Figure \ref{fig:non-iso_graphs}, and cited from \cite{huang2022boosting}),
while 3-WL cannot, which suggests that UnionSNN is stronger than 3-WL on such graphs. 
The induced graphs of arbitrary nodes $v$ and $v^\prime$ in the two graphs cannot be distinguished by 1-WL. Thus the original graphs cannot be distinguished by 3-WL. However, the numbers of 3-cycles and 4-cycles in their union subgraphs are different, and UnionSNN is able to reflect the number of 3-cycles and 4-cycles with the consideration of $E^{vu}_3$ and $E^{vu}_4$ edges in union subgraphs.
\jiaxing{In addition, the range of the union subgraph of an edge is inherently limited to 3-hop neighbors of each end node of the edge. As a result, one should expect that UnionSNN’s power is theoretically upper-bounded by 4-WL. Furthermore, UnionSNN has limitations in distinguishing cycles with length larger than 6, as this would require information further than 3 hops.} 

\section{Experimental Study}

In this section, we evaluate the effectiveness of our proposed model under various settings and aim to answer the following research questions: \textbf{RQ1.} Can UnionSNN outperform existing MPNNs and transformer-based models? \textbf{RQ2.} Can other GNNs benefit from our structural coefficient? \textbf{RQ3.} How do different components affect the performance of UnionSNN? \textbf{RQ4.} Is our runtime competitive with other substructure descriptors? We conduct experiments on four tasks: graph classification, graph regression, node classification and cycle detection. When we use UnionSNN to plugin other models we use the prefix term "Union-", such as UnionGCN. The implementation details are introduced in Appendix \ref{app:implementation}.

\noindent
\textbf{Datasets.} For graph classification, we use 10 benchmark datasets.  Eight of them were selected from the TUDataset \cite{KKMMN2016}, including MUTAG, PROTEINS, ENZYMES, DD, FRANKENSTEIN (denoted as FRANK in our tables), Tox21, NCI1 and NCI109. The other two datasets OGBG-MOLHIV and OGBG-MOLBBBP were selected from Open Graph Benchmark \cite{hu2020open}. For graph regression, we conduct experiments on ZINC10k and ZINC-full datasets \cite{dwivedi2020benchmarkgnns}. For node classification, we test on five datasets, including citation networks (Cora, Citeseer, and PubMed \cite{sen2008collective}) and Amazon co-purchase networks (Computer and Photo \cite{mcauley2015image}). For cycle detection, we conduct experiments on the detection of cycles of lengths 4 and 6, implemented as a graph classification task \cite{KKMMN2016}. These datasets cover various graph sizes and densities. 
Statistics of the datasets used are summarized in Tables \ref{tab:graph_dataset} and \ref{tab:node_dataset}.

\begin{table*}[ht]
\centering
\small
\setlength\tabcolsep{1.2pt}
\begin{tabular}{l|cccccccc}
\hline
          & MUTAG                         & PROTEINS                & ENZYMES                 & DD                      & FRANK                   & Tox21                  & NCI1     & NCI109  \\ \hline
GAT       & 77.56   \tiny{± 10.49}               & 74.34   \tiny{± 2.09}          &  67.67   \tiny{± 3.74}    & 74.25   \tiny{± 3.76}          & 62.85   \tiny{± 1.90}          & 90.35   \tiny{± 0.71}    & 78.07 \tiny{± 1.94}     & 74.34 \tiny{± 2.18}     \\
3WL-GNN   & 84.06   \tiny{± 6.62}                & 60.18   \tiny{± 6.35}          & 54.17   \tiny{± 6.25}          & 74.84   \tiny{± 2.63}    & 58.68   \tiny{± 1.93}          & 90.31   \tiny{± 1.33}   & 78.39 \tiny{± 1.54}     & 77.97 \tiny{± 2.22}     \\
UGformer  & 75.66   \tiny{± 8.67}         &  70.17   \tiny{± 5.42}    & 64.57   \tiny{± 4.53}          & 75.51   \tiny{± 3.52}          & 56.13   \tiny{± 2.51}          & 88.06   \tiny{± 0.50}     & 68.84 \tiny{± 1.54}     & 66.37 \tiny{± 2.74}       \\
MEWISPool & 84.73   \tiny{± 4.73}                & 68.10   \tiny{± 3.97}          &  53.66   \tiny{± 6.07}    &  76.03   \tiny{± 2.59}    & 64.63   \tiny{± 2.83}          & 88.13   \tiny{± 0.05}    & 74.21 \tiny{± 3.26}     & 75.30 \tiny{± 1.45}        \\
CurvGN    & 87.25   \tiny{± 6.28}     & 75.73   \tiny{± 2.87}    & 56.50   \tiny{± 7.13}   & 72.16   \tiny{± 1.88}   & 61.89   \tiny{± 2.41}   & 90.87   \tiny{± 0.38}   & 79.32 \tiny{± 1.65}     & 77.30 \tiny{± 1.78}  \\
NestedGIN    & 86.23   \tiny{± 8.82}     & 68.55   \tiny{± 3.22}    & 54.67   \tiny{± 9.99}   & 70.04   \tiny{± 4.32}   & 67.07   \tiny{± 1.46}   & 91.42   \tiny{± 1.18}   & 82.04 \tiny{± 2.23}    & 79.94 \tiny{± 1.59}  \\
DropGIN & 84.09 ± \tiny{8.42} & 73.39 ± \tiny{4.90} & 67.50 ± \tiny{6.42} & 71.05 ± \tiny{2.68} & 66.91 ± \tiny{2.16} & 91.66 ± \tiny{0.92} & 82.19 ± \tiny{1.39} & 81.13 ± \tiny{0.43} \\
GatedGCN-LSPE & \underline{88.33}   \tiny{± 3.88} & 73.94   \tiny{± 2.72} & 64.50   \tiny{± 5.92} & \underline{76.74}   \tiny{± 2.04} & 67.74   \tiny{± 2.65} & 91.71   \tiny{± 0.71} & 80.75   \tiny{± 1.67} & 80.13   \tiny{± 2.33} \\ 
GraphSNN  & 84.04   \tiny{± 4.09}                & 71.78   \tiny{± 4.11}          &  67.67   \tiny{± 3.74}    & 76.03   \tiny{± 2.59}    & 67.17   \tiny{± 2.25}          & \textbf{92.24}   \tiny{± 0.59}    & 70.87 \tiny{± 2.78}     & 70.11 \tiny{± 1.86}        \\ \hline
GCN       & 77.13 \tiny{± 5.24}     & 73.89   \tiny{± 2.85}          & 64.33   \tiny{± 5.83}          & \cellcolor{lightgray}{72.16}   \tiny{± 2.83}          & 58.80   \tiny{± 1.06}          & 90.10   \tiny{± 0.77}        & 79.73 \tiny{± 0.95}     & 75.91 \tiny{± 1.53} \\
UnionGCN (ours)   & \cellcolor{lightgray}{81.87} \tiny{± 3.81}       & \cellcolor{lightgray}{75.02}   \tiny{± 2.50}     & \cellcolor{lightgray}{64.67}   \tiny{± 7.14}          & 69.69   \tiny{± 4.18}          & \cellcolor{lightgray}{61.72}   \tiny{± 1.76}          & \cellcolor{lightgray}{91.63}   \tiny{± 0.72}     & \cellcolor{lightgray}{80.41} \tiny{± 1.94}     & \cellcolor{lightgray}{79.50} \tiny{± 1.82}    \\ \hline
GatedGCN  & 77.11   \tiny{± 10.05}               & \underline{76.18}  \tiny{± 3.12}    &  66.83   \tiny{± 5.08}    & \cellcolor{lightgray}{72.58}   \tiny{± 3.04}          & 61.40   \tiny{± 1.92}          & 90.83   \tiny{± 0.96}    & 80.32 \tiny{± 2.07}     & 78.19 \tiny{± 2.39} \\
UnionGatedGCN(ours)  & \cellcolor{lightgray}{77.14}   \tiny{± 8.14}               & \cellcolor{lightgray}{\textbf{76.91}}  \tiny{± 3.06}    &  \underline{\cellcolor{lightgray}{67.83}}   \tiny{± 6.87}    & 72.50   \tiny{± 2.22}          & \cellcolor{lightgray}{61.47}   \tiny{± 2.54}          & \cellcolor{lightgray}{91.31}   \tiny{± 0.89}    & \cellcolor{lightgray}{80.95} \tiny{± 2.11}     & \cellcolor{lightgray}{78.21} \tiny{± 2.58} \\ \hline
GraphSAGE & 80.38   \tiny{± 10.98}               & \cellcolor{lightgray}{74.87}   \tiny{± 3.38}          & 52.50   \tiny{± 5.69}          & 73.10   \tiny{± 4.34}          & 52.95   \tiny{± 4.01}          & 88.36   \tiny{± 0.15}    & 63.94 \tiny{± 2.52}     & 65.46 \tiny{± 1.12}    \\
UnionGraphSAGE(ours) & \cellcolor{lightgray}{83.04}   \tiny{± 8.70}               & 74.57   \tiny{± 2.35}          & \cellcolor{lightgray}{58.32}   \tiny{± 2.64}          & \cellcolor{lightgray}{73.85}   \tiny{± 4.46}          & \cellcolor{lightgray}{56.75}   \tiny{± 3.85}          & \cellcolor{lightgray}{88.59}   \tiny{± 0.12}    & \cellcolor{lightgray}{69.36} \tiny{± 1.64}     & \cellcolor{lightgray}{69.87} \tiny{± 1.04}    \\ \hline
GIN       & 86.23   \tiny{± 8.17}                & 72.86   \tiny{± 4.14}          & 65.83   \tiny{± 5.93}          & 70.29   \tiny{± 2.96}          & 66.50   \tiny{± 2.37}    &  \cellcolor{lightgray}{91.74}   \tiny{± 0.95}     & \underline{\cellcolor{lightgray}{82.29}} \tiny{± 1.77}     & 80.95 \tiny{± 1.87}  \\
UnionGIN (ours)   & \textbf{\cellcolor{lightgray}{88.86}}   \tiny{± 4.33}                & \cellcolor{lightgray}{73.22}   \tiny{± 3.90}          & \underline{\cellcolor{lightgray}{67.83}}   \tiny{± 6.10}          & \cellcolor{lightgray}{70.47}   \tiny{± 4.98}          & \textbf{\cellcolor{lightgray}{68.02}}   \tiny{± 1.47}          & \cellcolor{lightgray}{91.74}   \tiny{± 0.74}        & \underline{\cellcolor{lightgray}{82.29}}  \tiny{± 1.98}     & \textbf{\cellcolor{lightgray}{82.24}} \tiny{± 1.24}  \\ \hline
UnionSNN (ours)   & 87.31   \tiny{± 5.29}       & 75.02   \tiny{± 2.50}          & \textbf{68.17}   \tiny{± 5.70}   & \textbf{77.00}   \tiny{± 2.37}  & \underline{67.83}   \tiny{± 1.99} & \underline{91.76}   \tiny{± 0.85}   & \textbf{82.34} \tiny{± 1.93}     & \underline{81.61} \tiny{± 1.78} \\ \hline
\end{tabular}
\caption{Graph classification results (average accuracy ± standard deviation) over 10-fold-CV. The first and second best results on each dataset are highlighted in \textbf{bold} and \underline{underlined}. The winner between a base model with and without our structural coefficient injected is highlighted in \colorbox{lightgray}{gray background}.  The same applies to all tables.} 
\label{tab:graph_classification}
\end{table*}

\noindent
\textbf{Baseline Models.} We select various GNN models as baselines, including (1) classical MPNNs such as GCN \cite{kipf2016semi}, GIN \cite{xu2018powerful}, GraphSAGE \cite{hamilton2017inductive}, GAT \cite{velivckovic2017graph}, GatedGCN \cite{bresson2017residual}; 
(2) WL-based GNNs such as 3WL-GNN \cite{maron2019provably}; 
(3) transformer-based methods such as UGformer \cite{nguyen2019universal}, Graphormer \cite{ying2021graphormer} and GPS \cite{rampavsek2022recipe}; 
(4) state-of-the-art graph pooling methods such as MEWISPool \cite{nouranizadeh2021maximum}; 
(5) methods that introduce structural information by shortest paths or curvature, such as GeniePath \cite{liu2019geniepath}, CurvGN \cite{ye2019curvature}, and NestedGIN \cite{zhang2021nested}; 
(6) GNN with subgraph aggregation method, such as DropGIN \cite{papp2021dropgnn};
(7) GNNs with positional encoding, such as GatedGCN-LSPE \cite{dwivedi2021graph}; 
(8) GraphSNN \cite{wijesinghe2021new}. 

\begin{table}[h]
\centering
\scalebox{0.9}{
\begin{tabular}{l|cc}
\hline
model        & OGBG-MOLHIV    & OGBG-MOLBBBP \\ \hline
GraphSAGE    & 70.37   ± 0.42 & 60.78 ± 2.43 \\
GCN          & 73.49   ± 1.99 & 64.04 ± 0.43 \\
GIN          & 70.60 ± 2.56   & 64.10 ± 1.05 \\
GAT          & 70.60 ± 1.78   & 63.30 ± 0.53 \\
CurvGN       & 73.17 ± 0.89   & 66.51 ± 0.80 \\
GraphSNN     & 73.05 ± 0.40   & 62.84 ± 0.36 \\
UnionSNN (ours)      & \textbf{74.44} ± 1.21   & \textbf{68.28} ± 1.47 \\ \hline
\end{tabular}
}
\caption{Graph classification results (average ROC-AUC ± standard deviation) on OGBG-MOLHIV and OGBG-MOLBBBP datasets. The best result is highlighted in \textbf{bold}.}
\label{tab:ogb_results}
\end{table}

\subsection{Performance on Different Graph Tasks}

\noindent
\textbf{Graph-Level Tasks}. For graph classification, we report the results on 8 TUDatasets in Table \ref{tab:graph_classification} and the results on 2 OGB datasets in Table \ref{tab:ogb_results}. Our UnionSNN outperforms all baselines in 7 out of 10 datasets (by comparing UnionSNN with all baselines without ``ours''). We further apply our structural coefficient as a plugin component to four MPNNs: GCN, GatedGCN, GraphSAGE and GIN. The results show that our structural coefficient is able to boost the performance of the base model in almost all cases, with an improvement of up to 11.09\% \jiaxing{(Achieved when plugging in our local encoding into GraphSAGE on the ENZYMES dataset: (58.32\% - 52.50\%) / 52.50\% = 11.09\%.)}. 
For graph regression, we report the mean absolute error (MAE) on ZINC10k and ZINC-full. Table \ref{tab:graph_reg_exp} shows the performance of MPNNs with our structural coefficient (UnionGCN, UnionGIN, UnionGatedGCN and UnionGraphSAGE) dramatically beat their counterparts. Additionally, when injecting our structural coefficient into Transformer-based models, Unionormer and UnionGPS further improve Graphormer and GPS.

\begin{table}[h]
\small
\centering
\vspace{-0.3cm}
\begin{tabular}{l|cc}
\hline
               & ZINC10k & ZINC-full \\ \hline
GCN            & 0.3800 \tiny{± 0.0171} & 0.1152 \tiny{± 0.0010}    \\
UnionGCN (ours)       & \cellcolor{lightgray}{0.2811} \tiny{± 0.0050}  & \cellcolor{lightgray}{0.0877 \tiny{  ± 0.0003}}    \\ \hline
GIN            & 0.5260 \tiny{± 0.0365}  & 0.1552 \tiny{± 0.0079}    \\
UnionGIN  (ours)     & \cellcolor{lightgray}{0.4625} \tiny{± 0.0222}  & \cellcolor{lightgray}{0.1334 \tiny{± 0.0013}}    \\ \hline
GraphSAGE      & 0.3980 \tiny{± 0.0290}  & 0.1205 \tiny{± 0.0034}    \\
UnionSAGE (ours) & \cellcolor{lightgray}{0.3768} \tiny{± 0.0041} & \cellcolor{lightgray}{0.1146 \tiny{± 0.0017}}    \\ \hline
Graphormer     & 0.1269 \tiny{± 0.0083} & 0.0309 \tiny{± 0.0031}    \\
Unionormer (ours)    & \cellcolor{lightgray}{0.1241} \tiny{± 0.0056}  & \cellcolor{lightgray}{0.0252} \tiny{± 0.0026}    \\ \hline
GPS     & 0.0740 \tiny{± 0.0022} & 0.0262 \tiny{± 0.0025}    \\
UnionGPS (ours)    & \cellcolor{lightgray}{\textbf{0.0681} \tiny{± 0.0013}}  & \cellcolor{lightgray}\textbf{0.0236} \tiny{± 0.0017}    \\ \hline
\end{tabular}
\caption{Graph regression results (average test MAE ± standard deviation) on ZINC10k and ZINC-full datasets.}
\label{tab:graph_reg_exp}
\end{table}

\begin{table}[h]
\centering
\scalebox{0.8}{
\begin{tabular}{l|cc}
\hline
Model             & 4-CYCLES & 6-CYCLES \\ \hline
GraphSAGE         & 50.00    & 50.00    \\
CGN               & 84.33    & 63.56    \\
GraphSNN          & 67.10    & 66.04    \\ \hline
GCN               & 50.00    & 50.00    \\
UnionGCN   (ours) & \cellcolor{lightgray}{94.31}    & \cellcolor{lightgray}{\textbf{95.59}}    \\ \hline
GIN               & 96.61    & 90.06    \\
UnionGIN   (ours) & \cellcolor{lightgray}{97.15}    & \cellcolor{lightgray}{92.63}    \\ \hline
UnionSNN   (ours) & \textbf{99.11}    & 95.00    \\ \hline
\end{tabular}
}
\caption{Cycle detection results (average accuracy ± standard deviation).}
\label{tab:cycles}
\end{table}

\begin{table*}[h!]
\centering
\scalebox{1.0}{
\begin{tabular}{l|cccccc}
\hline
          & Cora           & Citeseer       & PubMed   & Computer     & Photo \\ \hline
GraphSAGE & 70.60 ±   0.64 & 55.02 ±   3.40 & 70.36 ±   4.29 & 80.30 ± 1.30 & 89.16  ± 1.03 \\
GAT       & 74.82 ±   1.95 & 63.82 ±   2.81 & 74.02 ±   1.11 & 85.94 ± 2.35 & 91.86  ± 0.47 \\
GeniePath & 72.16 ±   2.69 & 57.40 ±   2.16 & 70.96 ±   2.06 & 82.68 ± 0.45 & 89.98  ± 1.14 \\
CurvGN    & 74.06 ±   1.54 & 62.08 ±   0.85 & 74.54 ±   1.61 & 86.30 ± 0.70 & 92.50  ± 0.50 \\ \hline
GCN       & 72.56 ±   4.41 & 58.30 ±   6.32 & 74.44 ±   0.71  & 84.58 ± 3.02 & 91.71  ± 0.55
 \\
UnionGCN (ours)   &   \cellcolor{lightgray}{74.48} ±   0.42 &  \cellcolor{lightgray}{59.02} ±   3.64 & \cellcolor{lightgray}{74.82} ±   1.10 & \textbf{\cellcolor{lightgray}{88.84}} ± 0.27 & \underline{\cellcolor{lightgray}{92.33}}  ± 0.53\\ \hline
GIN       & 75.86 ±   1.09 & 63.10 ±   2.24 & 76.62 ±   0.64 & 86.26 ± 0.56 & 92.11  ± 0.32 \\
UnionGIN (ours)   & \underline{\cellcolor{lightgray}{75.90}} ±   0.80 & \cellcolor{lightgray}{63.66} ±   1.75 & \cellcolor{lightgray}{76.78} ±   1.02 & \cellcolor{lightgray}{86.81} ± 2.12 & \cellcolor{lightgray}{92.28}  ± 0.19\\ \hline
GraphSNN  & 75.44 ± 0.73    & 64.68 ± 2.72   &   76.76 ± 0.54  & 84.11 ± 0.57 & \cellcolor{lightgray}{90.82}  ± 0.30           \\
UnionGraphSNN (ours)  & \cellcolor{lightgray}{75.58} ± 0.49    & \cellcolor{lightgray}{\textbf{65.22}} ± 1.12   &   \underline{\cellcolor{lightgray}{76.92}} ± 0.56   &  \cellcolor{lightgray}{84.58} ± 0.46 & 90.60  ± 0.58         \\ \hline
UnionSNN (ours)   & \textbf{76.86} ± 1.58   & \underline{65.02} ± 1.02   & \textbf{77.06} ± 1.07  & \underline{87.76} ± 0.36 & \textbf{92.92}  ± 0.38 \\ \hline
\end{tabular}
}
\caption{Node classification results (average accuracy ± standard deviation) over 10 runs.}
\label{tab:node_classification}
\end{table*}

\begin{table*}[h!]
\centering
\small
\begin{tabular}{l|cccccc}
\hline
                         & MUTAG          & PROTEINS       & ENZYMES        & DD             & NCI1           & NCI109         \\ \hline
overlap         & 85.70 ±   7.40 & 71.33 ±   5.35 & 65.00 ±   5.63 & 73.43 ±   4.07 & 73.58 ±   1.73 & 72.96 ±   2.01 \\
minus & \textbf{87.31} ±   5.29 & 68.70 ±   3.61 & 65.33 ±   4.58 & 74.79 ± 4.63 & 80.66 ±   1.90 & 78.70 ±   2.48 \\
union                  & \textbf{87.31} ±   5.29 & \textbf{75.02} ± 2.50   & \textbf{68.17} ±   5.70 & \textbf{77.00} ± 2.37 & \textbf{82.34} ±   1.93 & \textbf{81.61} ±   1.78 \\ \hline
\end{tabular}
\caption{Ablation study on local substructure.}
\label{tab:ablation_1}
\end{table*}

\begin{table*}[h!]
\centering
\small
\begin{tabular}{l|cccccc}
\hline
             & MUTAG          & PROTEINS       & ENZYMES        & DD             & NCI1           & NCI109         \\ \hline
BetSNN       & 80.94 ±   6.60 & 69.44 ±   6.15 & 65.00 ±   5.63 & 70.20 ±   5.15 & 74.91 ±   2.48 & 73.70 ±   1.87 \\
CountSNN     & 84.65 ±   6.76 & 70.79 ±   5.07 & 66.50 ±   6.77 & 74.36 ±   7.21 & 81.74 ±   2.35 & 79.80 ±   1.67 \\
CurvSNN & 85.15 ±   7.35 & 72.77 ±   4.42 & 67.17 ±   6.54 & 75.88 ±   3.24 & 81.34 ±   2.27 & 80.64 ±   1.85 \\
LapSNN       & \textbf{89.39} ±   5.24 & 68.32 ±   3.49 & 66.17 ±   4.15 & 76.31 ±   2.85 & 81.39 ±   2.08 & 81.34 ±   2.93 \\
UnionSNN      & 87.31 ±   5.29 & \textbf{75.02} ± 2.50   & \textbf{68.17} ±   5.70 & \textbf{77.00}  ± 2.37 & \textbf{82.34} ±   1.93 & \textbf{81.61} ±   1.78 \\ \hline
\end{tabular}
\caption{Ablation study on substructure descriptor.}
\label{tab:ablation_2}
\end{table*}

\begin{table*}[h!]
\centering
\small
\begin{tabular}{l|cccccc}
\hline
                    & MUTAG          & PROTEINS       & ENZYMES        & DD             & NCI1           & NCI109         \\ \hline
matrix sum        & \textbf{88.89} ±   7.19 & 71.32 ±   5.48 & 65.17 ±   6.43 & 70.71 ±   4.07 & 80.37 ±   2.73 & 79.84 ±   1.89 \\
eigen max & 86.73 ±   5.84 & 71.78 ±   3.24 & 67.67 ±   6.88 & 74.29 ±   3.26 & 81.37 ±   2.08 & 79.23 ±   2.01 \\
svd sum             & 87.31 ±   5.29 & \textbf{75.02} ± 2.50   & \textbf{68.17} ±   5.70 & \textbf{77.00} ±   2.37 & \textbf{82.34} ±   1.93 & \textbf{81.61} ±   1.78 \\ \hline
\end{tabular}
\caption{Ablation study on path matrix encoding method.}
\label{tab:ablation_3}
\end{table*}

Detecting cycles requires more than 1-WL expressivity \cite{morris2019weisfeiler}. To further demonstrate the effectiveness of UnionSNN, we conduct experiments on the detection of cycles of lengths 4 and 6. Table \ref{tab:cycles} shows the superiority of UnionSNN over 5 baseline models for detecting cycles. Note that GCN makes a random guess on the cycle detection with an accuracy of 50\%. By incorporating our structural coefficient to GCN, we are able to remarkably boost the accuracy to around 95\%. The large gap proves that our structural coefficient is highly effective in capturing local information. 

\noindent
\textbf{Node-Level Tasks}. We report the results of node classification in Table \ref{tab:node_classification}. UnionSNN outperforms all baselines on all 5 datasets. Again, injecting our structural coefficients to GCN, GIN, and GraphSNN achieves performance improvement over base models in almost all cases. \jiaxing{Remarkably, we find that integrating our structural coefficients into GCN, GIN, and GraphSNN can effectively reduce the standard deviations (stds) of the classification accuracy of these models in most cases, and some are with a large margin. For instance, the std of classification accuracy of GCN on Cora is reduced from 4.41 to 0.42 after injecting our structural coefficients.} 

\subsection{Ablation Study}

In this subsection, we validate empirically the design choices made in different components of our model: (1) the local substructure; (2) the substructure descriptor; (3) the encoding method from a path matrix to a scalar. All experiments were conducted on 6 graph classification datasets.

\noindent
\textbf{Local Substructure}. We test three types of local substructures defined in Section \ref{sec:structural}: overlap subgraphs, union minus subgraphs and union subgraphs. They are denoted as ``overlap'', ``minus'', and ``union'' respectively in Table \ref{tab:ablation_1}. The best results are consistently achieved by using union subgraphs.  

\noindent
\textbf{Substructure Descriptor}. We compare our substructure descriptor with four existing ones discussed in Section \ref{sec:descriptor}. We replace the substructure descriptor in UnionSNN with edge betweenness, node/edge counting,  Ricci curvature, and Laplacian matrix (other components unchanged), and obtain four variants, namely BetSNN, CountSNN, CurvSNN, and LapSNN. Table \ref{tab:ablation_2} shows our UnionSNN is a clear winner: it achieves the best result on 5 out of 6 datasets. 
This experiment demonstrates that our path matrix better captures structural information.

\noindent
\textbf{Path Matrix Encoding Method}. We test three methods that transform a path matrix to a scalar: (1) sum of all elements in the path matrix (matrix sum); (2) maximum eigenvalue of the path matrix (eigen max); (3) sum of all singular values of the matrix (svd sum) used by UnionSNN in Section \ref{sec:network}. Table \ref{tab:ablation_3} shows that the encoding method ``svd sum'' performs the best on 5 out of 6 datasets. 

\subsection{Case Study}

\begin{figure}[h]
\centering
\includegraphics[width=.9\linewidth]{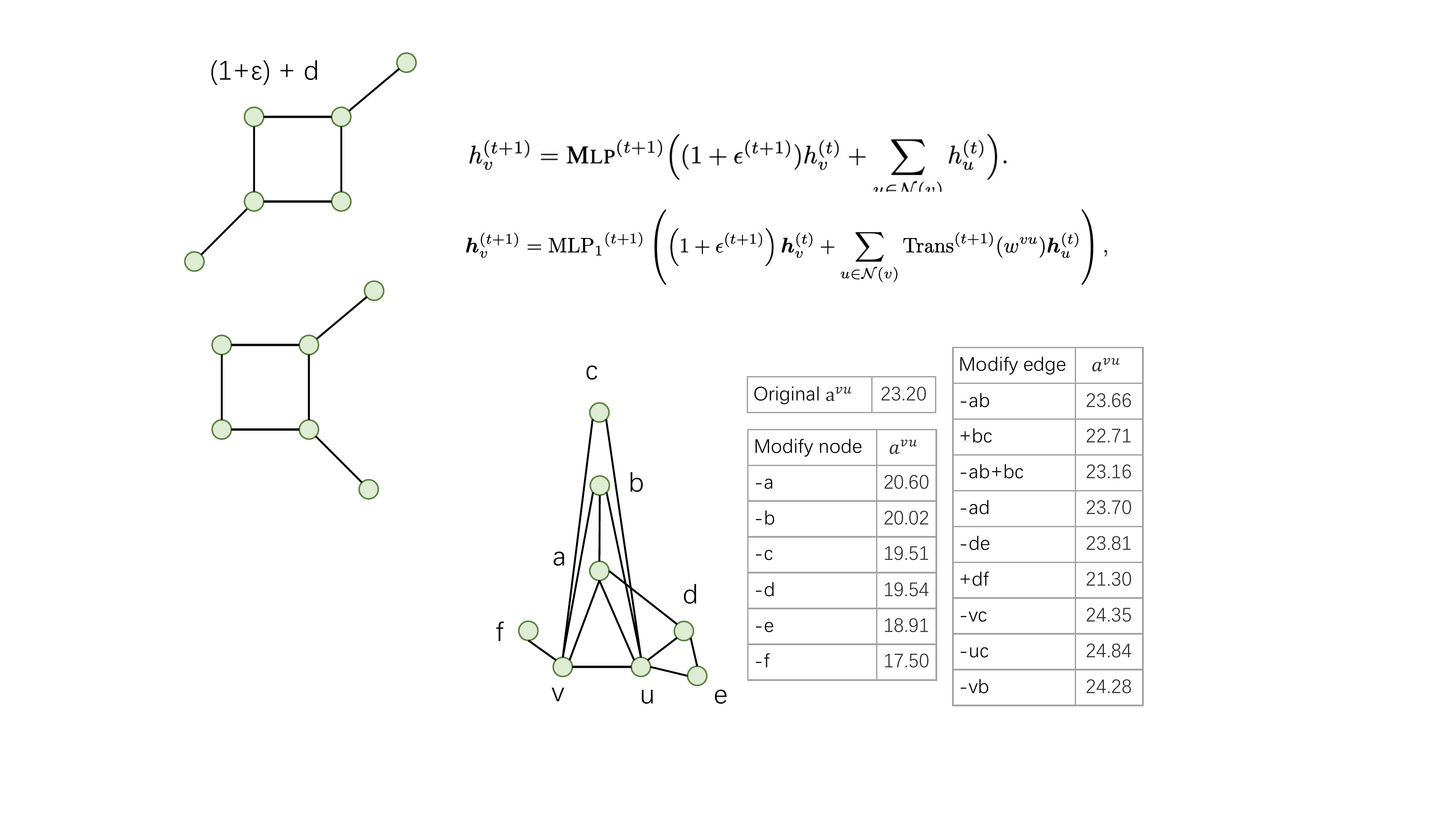}
\caption{Structural coefficient analysis.}
\label{fig:case_study}
\end{figure}

In this subsection, we investigate how the proposed structural coefficient $a^{vu}$ reflects local connectivities. We work on an example union subgraph $S_{v \cup u}$ in Figure \ref{fig:case_study} and modify its nodes/edges to study how the coefficient $a^{vu}$ varies with the local structural change.
We have the following observations: (1) with the set of nodes unchanged, deleting an edge increases $a^{vu}$; (2) deleting a node (and its incident edges) decreases $a^{vu}$; (3) the four types of edges in the closed neighborhood (Section \ref{sec:structural}) have different effects to $a^{vu}$: $E_1^{vu}$ \textless $E_2^{vu}$ \textless $E_3^{vu}$ \textless $E_4^{vu}$ (by comparing -ab, -ad, -de, and +df). These observations indicate that a smaller coefficient will be assigned to an edge with a denser local substructure. This matches our expectation that the coefficient should be small for an edge in a highly connected neighborhood. The rationale is, such edges are less important in message passing as the information between their two incident nodes can flow through more paths. By using the coefficients that well capture local connectivities, the messages from different neighbors could be properly adjusted when passing to the center node. This also explains the effectiveness of UnionSNN in performance experiments. 

\subsection{Efficiency Analysis}
\label{sec:eff_exp}

In this subsection, we conduct experiments on PROTEINS, DD and FRANKENSTEIN datasets, which cover various number of graphs and graph sizes. 

\noindent
\textbf{Preprocessing Computational Cost.} UnionSNN computes structural coefficients in preprocessing. We compare its preprocessing time with the time needed in baseline models for pre-computing their substructure descriptors, including edge betweenness (Betweenness) in BetSNN, node/edge counting (Count\_ne) in GraphSNN, Ricci curvature (Curvature) in CurvGN, and counting cycles (Count\_cycle) in \cite{bodnar2021weisfeiler}. As shown in Table \ref{tab:preprocess}, the preprocessing time of UnionSNN is comparable to that of other models. This demonstrates that our proposed structural coefficient 
is able to improve performance without significantly sacrificing efficiency. 
\jiaxing{Our computational cost could be further reduced by pre-computing in the input graph all node pairs with a distance of 1, 2, or 3 (all possible distances in our union subgraphs). The local path matrix could be computed efficiently by simple checking. This can avoid repeated shortest path computations in each local neighborhood.}

\begin{table}[h]
\centering
\small
\setlength\tabcolsep{2pt}
\begin{tabular}{l|ccc}
\hline
             & PROTEINS & DD  & FRANK \\ \hline
Betweenness  & 51       & 749   & 47    \\
Count\_ne    & 44       & 669   & 41    \\
Curvature    & 304      & 1295  & 1442  \\
Count\_cycle & 2286     & -    & 86385  \\
Ours    & 73       & 1226  & 59    \\ \hline
\end{tabular}
\caption{Time cost (seconds) for computing structural coefficients in preprocessing. We are not reporting the time of counting cycles on DD since it took more than 100 hours.}
\label{tab:preprocess}
\end{table}

\noindent
\textbf{Runtime Computational Cost} We conduct an experiment to compare the total runtime cost of UnionSNN with those in other MPNNs. The results are reported in Table \ref{tab:total_time}. Although UnionSNN runs slightly slower than GCN and GIN, it runs over 4.56 times faster than WL-based MPNN (3WL-GNN) and is comparable to MPNN with positional encoding (GatedGCN-LSPE). Compared with GraphSNN, UnionSNN runs significantly faster: the efficiency improvement approaches an order of magnitude on datasets with large graphs, e.g., DD. This is because UnionSNN does not need to pad the adjacency matrix and the feature matrix of each graph to the maximum graph size in the dataset, as GraphSNN does. 

\begin{table}[h]
\centering
\footnotesize
\begin{tabular}{l|ccc}
\hline
        & PROTEINS & DD & FRANK \\ \hline
GCN      & 0.67     & 1.51  & 1.65  \\
GIN      & 0.53     & 1.81  & 2.01  \\
3WL-GNN   & 6.06     & 75.25  & 19.31  \\
GatedGCN-LSPE & 1.33 & 3.65 & 2.55 \\
GraphSNN & 4.05     & 30.45 & 5.76  \\
UnionSNN  & 1.31     & 3.61  & 2.46  \\ \hline
\end{tabular}
\caption{Time cost (hours) for a single run with 10-fold-CV, including training, validation, test (excluding preprocessing).}
\label{tab:total_time}
\end{table}

\section{Conclusions}

We present UnionSNN, a model that outperforms 1-WL in distinguishing non-isomorphic graphs. UnionSNN utilizes an effective shortest-path-based substructure descriptor applied to union subgraphs, making it more powerful than previous models. Our experiments demonstrate that UnionSNN surpasses state-of-the-art baselines in both graph-level and node-level classification tasks while maintaining its computational efficiency. The use of union subgraphs enhances the model's ability to capture neighbor connectivities and facilitates message passing. Additionally, when applied to existing MPNNs and Transformer-based models, UnionSNN improves their performance by up to 11.09\%.

\newpage

\section*{Acknowledgments}
\jiaxing{This research/project is supported by the National Research Foundation, Singapore under its Industry Alignment Fund – Pre-positioning (IAF-PP) Funding Initiative, and the Ministry of Education, Singapore under its MOE Academic Research Fund Tier 2 (STEM RIE2025 Award MOE-T2EP20220-0006). Any opinions, findings and conclusions or recommendations expressed in this material are those of the author(s) and do not reflect the views of National Research Foundation, Singapore, and the Ministry of Education, Singapore.}

\bibliography{aaai24}

\clearpage

\appendix

\section{An Example of the Power of Local Substructures}
\label{app:substructures}

\begin{figure*}[btp]
\begin{center}
\includegraphics[width=.8\linewidth]{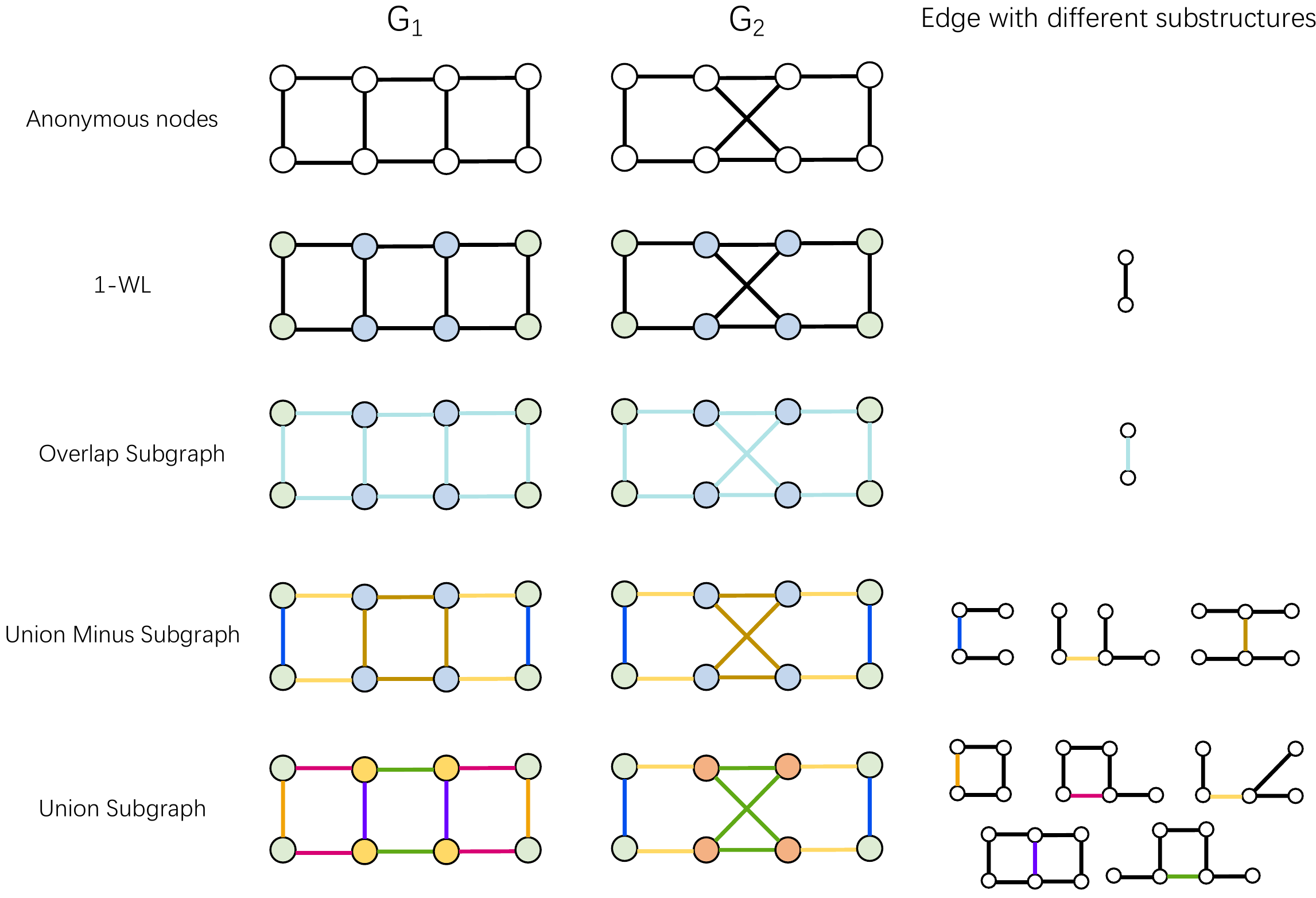}
\end{center}
\caption{An example graph pair and their color refinements by using different local substructures.}
\label{fig:substructures}
\end{figure*}

In Figure \ref{fig:substructures}, two graphs $G_1$ and $G_2$ contain different numbers of four cycles and are hence non-isomorphic. Since the refined colors by the 1-WL algorithm for these two graphs are the same, MPNNs will generate the same representation for them as well (see the colors generated by 1-WL and MPNNs in the second row). When using the overlap subgraph to encode the local substructure for edges, as shown in the third row, all edges will obtain the same color since their overlap subgraphs are the same. When applying the union minus subgraph for edge encoding, it detects different types of edges with various local substructures, but it is still not powerful enough to distinguish $G_1$ from $G_2$. Since the edges of the corresponding positions in the two graphs have the same color, the assigned colors of the nodes at the corresponding positions are also the same (see the fourth row). In contrast, when applying the union subgraph (the last row), edges in $G_1$ and $G_2$ clearly depict different local substructures, and thus the nodes and edges of the two graphs exhibit different colors. This example demonstrates that the union subgraph is a more powerful local substructure to distinguish non-isomorphic graph pairs when compared against 1-WL or other types of local substructures.

\section{Proofs of Theorems}
\label{app:proofs}

\subsection{Proof of Theorem 1}
\label{app:thm1}

\textbf{Theorem 1.} \textit{If $S_i \simeq_{union} S_j$, then $S_i \simeq_{overlap} S_j$; but not vice versa.}

\textit{Proof.} By the definition of union isomorphism, if $S_i \simeq_{union} S_j$, then there exists a bijective mapping  $g: \tilde{\mathcal{N}}(i) \rightarrow \tilde{\mathcal{N}}(j)$ such that $g(i) = j$ and for any $v \in \mathcal{N}(i)$ and $g(v) = u$, $S_{i \cup v}$ and $S_{j \cup u}$ are isomorphic. 
Since $g(i) = j$ and $g(v) = u$, according to the definition of graph isomorphism, $v_1 \in \tilde{\mathcal{N}}(i) \cap \tilde{\mathcal{N}}(v)$ if and only if $g(v_1) \in \tilde{\mathcal{N}}(j) \cap \tilde{\mathcal{N}}(u)$. Therefore, $g(\cdot)$ is a bijective mapping from $\tilde{\mathcal{N}}(i) \cap \tilde{\mathcal{N}}(v)$ to $\tilde{\mathcal{N}}(j) \cap \tilde{\mathcal{N}}(u)$, and $v_1, v_2 \in \tilde{\mathcal{N}}(i) \cap \tilde{\mathcal{N}}(v)$ are adjacent if and only if $g(v_1), g(v_2) \in \tilde{\mathcal{N}}(j) \cap \tilde{\mathcal{N}}(u)$ are adjacent. This proves that $S_{i \cap v}$ and $S_{j \cap u}$ are isomorphic, and thus $S_i \simeq_{overlap} S_j$. On the contrary, it is possible that $S_i \simeq_{overlap} S_j$ but $S_i \not \simeq_{union} S_j$, as shown by graphs $G_1$ and $G_2$ in Figure~\ref{fig:isoexample}. $\Box$

\subsection{Proof of Theorem 2}
\label{app:thm2}

\textbf{Theorem 2.} \textit{With a fixed order of nodes in the path matrix, we can obtain a unique path matrix $\mP^{vu}$ for a given union subgraph $S_{v \cup u}$, and vice versa.}


\textit{Proof.} We prove the theorem in two steps.

Step 1: We prove that we get a unique $\mP^{vu}$ from a given $S_{v \cup u}$. With the node order fixed, the rows and the columns of $\mP^{vu}$ are fixed. Since $\mP^{vu}$ stores the lengths of all-pair shortest paths, the matrix is unique given an input union subgraph. 

Step 2: We prove that we can recover a unique $S_{v \cup u}$ from a given $\mP^{vu}$. The node set of $S_{v \cup u}$ can be recovered from the row (or column) indices of $\mP^{vu}$. The edge set of $S_{v \cup u}$ can be recovered from the entries in $\mP^{vu}$ with the value of "1". Both the node set and the edge set can be uniquely constructed from $\mP^{vu}$, and thus $S_{v \cup u}$ is unique. $\Box$

\subsection{Proof of Theorem 3}
\label{app:thm3}


The proof of Theorem 3 follows a similar flow to that of Theorem 4 in \cite{wijesinghe2021new}. 

We first define the concept of subtree isomorphism. Let $h_v$ denote the node feature of a node $v \in V$.

\textbf{Subtree Isomorphism.} $S_i$ and $S_j$ are subtree-isomorphic, denoted as $S_i \simeq_{subtree} S_j$, if there exists a bijective mapping $g$: $\tilde{\mathcal{N}}(i) \rightarrow \tilde{\mathcal{N}}(j)$ such that $g(i) = j$, and for any $v \in \mathcal{N}(i)$ and $g(v) = u$, $h_{v}=h_{u}$.


We assume that $\mathcal{H},\mathcal{A}$ and $\mathcal{W}$ are three countable sets that $\mathcal{H}$ is the node feature space, $\mathcal{A}$ is the structural coefficient space, and $\mathcal{W} = \{ a^{ij} h_i | a^{ij} \in \mathcal{A}, h_i \in \mathcal{H} \}$. Suppose $H$ and $W$ are two multisets that contain elements from $\mathcal{H}$ and $\mathcal{W}$, respectively, and $|H| = |W|$. In order to prove Theorem 3, we need to use Lemmas 1 and 2 in \cite{wijesinghe2021new}. To be self-contained, we repeat the lemmas here and refer the readers to Appendix C of \cite{wijesinghe2021new} for the proofs.

\textbf{Lemma 1.} \textit{There exists a function $f$ s.t. $\pi (H,W) = \sum_{h \in H, w \in W} f(h,w)$ is unique for any distinct pair of multisets $(H,W)$.}



\textbf{Lemma 2.} \textit{There exists a function $f$ s.t. $\pi^\prime(h_v,H,W)=\gamma f(h_v, |H|h_v)+\sum_{h \in H, w \in W} f(h,w)$ is unique for any distinct $(h_v,H,W)$, where $h_v \in H, |H|h_v \in W$, and $\gamma$ can be an irrational number.}




    
    
    
    


From the lemmas above, we can now prove Theorem 3:

\textbf{Theorem 3.} \textit{UnionSNN is more expressive than 1-WL in testing non-isomorphic graphs.}

\textit{Proof.} By Theorem 3 in \cite{wijesinghe2021new}, if a GNN $M$ satisfies the two following conditions, then M is strictly more powerful than 1-WL in distinguishing non-isomorphic graphs.

\begin{enumerate}
    \item M can distinguish at least one pair of neighborhood subgraphs $S_i$ and $S_j$ such that $S_i$ and $S_j$ are subtree-isomorphic, but they are not isomorphic, and $\{\{ \tilde{a}^{iv} | v \in \mathcal{N}(i)\}\} \neq \{\{ \tilde{a}^{ju} | u \in \mathcal{N}(j) \}\}$, where $\tilde{a}^{vu}$ is the normalized value of $a^{vu}$;
    \item The aggregation scheme $\Phi(h_v^{(t)}, \{\{h_u^{(t)} | u \in \mathcal{N}(v) \}\}, \{\{\ (\tilde{a}^{uv},h_u^{(t)}) | u \in \mathcal{N}(v) \}\}) $ is injective.
\end{enumerate}

For condition 1, the pair of graphs in Figure~\ref{fig:isoexample} satisfies the condition, and can be distinguished by UnionSNN as they are not union-isomorphic.

For condition 2, by Lemmas 1 and 2 and the fact that the MLP $\operatorname{Trans}(\cdot)$ is a universal approximator and can model and learn the required functions, we conclude that UnionSNN satisfies this condition.

Therefore, UnionSNN is more expressive than 1-WL in testing non-isomorphic graphs. $\Box$

\section{Definitions of Three Other Substructure Descriptor Functions}
\label{app:sub_des_func}

\textbf{Edge Betweenness.} The betweenness centrality of an edge $e \in E$ in a graph $G = (V, E, X)$ is defined as the sum of the fraction of all-pair shortest paths that pass through $e$:

\begin{equation}
c_G(e)=\sum_{v, u \in V} \frac{\sigma(v, u, G|e)}{\sigma(v, u, G)},
\label{eq:betweenness}
\end{equation}

\noindent
where $\sigma(v, u, G)$ is the number of shortest paths between $v$ and $u$, and $\sigma(v, u, G|e)$ is the number of those paths passing through edge $e$.

\noindent
\textbf{Node/Edge Counting.} GraphSNN \cite{wijesinghe2021new} defines a structural coefficient based on the number of nodes and edges in the (sub)graph. When applied to the union subgraph, we have

\begin{equation}
\omega(S_{v \cup u}) = \frac{|E_{v \cup u}|}{|V_{v \cup u}| \cdot |V_{v \cup u} - 1|}|V_{v \cup u}|^\lambda,
\label{eq:count}
\end{equation}

\noindent
where $\lambda=1$ for node classification and $\lambda=2$ for graph classification.

\noindent
\textbf{Ollivier Ricci Curvature.} Given a node $v \in V$ in a graph $G = (V, E, X)$, a probability vector of $v$ is defined as:

\begin{equation}
\mu_v^\alpha: u \mapsto\left\{\begin{array}{l}
\alpha, u=v \\
\frac{1-\alpha}{d_v}, u \in \mathcal{N}(v), \\
0, \text { otherwise }
\end{array}\right.
\label{eq:distribution}
\end{equation}

\noindent
where $\alpha \in [0,1)$ is a hyperparameter. Following \cite{ye2019curvature}, we use $\alpha=0.5$ in all our experiments. The $\alpha$-Ricci curvature $\kappa_{vu}^\alpha$ on an edge $(v, u) \in E$ is defined by:

\begin{equation}
\kappa_{vu}^\alpha = 1 - \frac{\operatorname{Wass}(\mu_v^\alpha, \mu_u^\alpha)}{d(v, u)},
\label{eq:curvature}
\end{equation}

\noindent
where $\operatorname{Wass}(\cdot)$ denotes the Wasserstein distance and $d(\cdot)$ denotes the shortest path length between $v$ and $u$. The Wasserstein distance can be estimated by the optimal transportation distance, which can be solved by the following linear programming:

\begin{equation}
\begin{split}
\min_M \sum_{i \in \tilde{\mathcal{N}}(v), j \in \tilde{\mathcal{N}}(u)} d\left(i, j\right) M\left(i, j\right) \\
\text { s.t. } \sum_{j \in \tilde{\mathcal{N}}(u)} M\left(i, j\right)=\mu_v^\alpha\left(i\right), \forall  i \in {\tilde{\mathcal{N}}}(v);\\
\sum_{i \in \tilde{\mathcal{N}}(v)} M\left(i, j\right)=\mu_u^\alpha\left(j\right), \forall j \in {\tilde{\mathcal{N}}}(u).
\label{eq:otd}
\end{split}
\end{equation}












\section{Algorithm of UnionSNN}
\label{app:code}

See Algorithm \ref{alg:unionsnn} in the following.

\begin{algorithm}[h]
	\caption{Algorithm for an $L$ layer UnionSNN.}
	\label{alg:unionsnn}
	
    \begin{flushleft}
    \noindent
	\textbf{Input}: A graph $G = (V, E, X)$ with $|V|$ nodes and $|E|$ edges; The UnionSNN update function as in Eq. (\ref{eq:UnionSNN}); A shortest path substructure descriptor $\operatorname{PathLen}$, as in Eq. (\ref{eq:path_matrix}); A feature transformation function $\operatorname{Trans}$, as in Eq.(\ref{eq:mlp2}). \\
  
    \noindent
    \textbf{Output}: Node representations $H^{(L)} \in \mathbb{R}^{N \times D}$ that can be fed into a downstream prediction head for graph/node-level prediction.
    \end{flushleft}
    \begin{algorithmic}[1]
            \FOR{Edge $(v, u) \in E$} 
		\STATE $\mP_{ij}^{vu} = \operatorname{PathLen}(i, j, S_{v \cup u}), i, j \in V_{v\cup u}$ \hfill \textsl{\color{teal!70} \# Compute shortest path matrix for each edge}
            \STATE $\mP=\mU \boldsymbol{\Sigma} \mV^*$ \hfill \textsl{\color{teal!70} \# Compute the singular values}
            \STATE $a^{vu} = \operatorname{sum}(\boldsymbol{\Sigma}^{vu})$ \hfill \textsl{\color{teal!70} \# Compute structural coefficient for each edge}
            \STATE $\tilde{a}^{vu} = \frac{a^{vu}}{\sum_{u \in \mathcal{N}(v)}a^{vu}}$
            \ENDFOR

		\FOR{$l = 1, 2, ... , L$}
            \STATE Update node representations by Eq. (\ref{eq:UnionSNN})
		
		\ENDFOR

            \STATE return $H^{(L)}$
	\end{algorithmic}
\end{algorithm}

\section{Implementation Details}
\label{app:implementation}

\textbf{TU Datasets.} We following the setup of \cite{dwivedi2020benchmarkgnns}, which splits each dataset into 8:1:1 for training, validation and test, respectively. The model evaluation and selection are done by collecting the accuracy from the single epoch with the best 10-fold cross-validation averaged accuracy. We choose the best values of the initialized learning rate from \{0.02, 0.01, 0.005, 0.001\}, weight decay from \{0.005, 0.001, 0.0005\}, hidden dimension from \{64, 128, 256\} and dropout from \{0, 0.2, 0.4\}.

\noindent
\textbf{OGB Datasets.} For graph classification on OGB datasets, we use the scaffold splits, run 10 times with different random seeds, and report the average ROC-AUC due to the class imbalance issue. We fix the initialized learning rate to 0.001, weight decay to 0, and dropout to 0. We keep the total number of parameters of all GNNs to be less than 100K by adjusting the hidden dimensions.

\noindent
\textbf{ZINC Datasets.} For graph regression, we use the data split of ZINC10k and ZINC-full datasets from \cite{dwivedi2020benchmarkgnns}, run 5 times with different random seeds, and report the average test MAE. The edge features have not been used. We fix the initialized learning rate to 0.001, weight decay to 0, and dropout to 0. By adjusting the hidden dimensions, we keep the total number of parameters of all GNNs to be less than 500K. For the GPS model, we choose GINE \cite{hu2019strategies} as the MPNN layer, Graphormer as the GlobAttn layer and the random-walk structural encoding (RWSE) as the positional encoding.

\noindent
\textbf{Cycle Detection Datasets.}  The datasets are proposed by \cite{vignac2020building}, in which graphs are labeled by containing 4/6 cycles or not. Each dataset contains 9000 graphs for training, 1000 graphs for validation and 10000 graphs for test. The settings of models are the same with OGB datasets.

\noindent
\textbf{Node Classification Datasets.} For semi-supervised node classification, we use the standard split in the original paper and report the average accuracy for 10 runs with different random seeds. We use a 0.001 learning rate with 0 dropout, and choose the best values of the weight decay from \{0.005, 0.001, 0.0005\} and hidden dimension from \{64, 128, 256\}.


For all experiments, the whole network is trained in an end-to-end manner using the Adam optimizer \cite{kingma2014adam}. We use the early stopping criterion, i.e., we stop the training once there is no further improvement on the validation loss during 25 epochs. The implementation of Graphormer and GPS is referred to their official code\footnote{\url{https://github.com/rampasek/GraphGPS}}, which is built on top of PyG \cite{fey2019fast} and GraphGym \cite{you2020design}. The code of all the other methods was implemented using PyTorch \cite{paszke2017automatic} and Deep Graph Library \cite{wang2019deep} packages. All experiments were conducted in a Linux server with Intel(R) Core(TM) i9-10940X CPU (3.30GHz), GeForce GTX 3090 GPU, and 125GB RAM.

\section{Time Complexity of Structure Descriptors}
\label{app:time}

Given a graph $G = (V, E)$, where $|V| = N$ and $|E| = M$, assume $n$ and $m$ are the average node number and edge number among all union subgraphs in the whole graph, respectively. The complexity of each substructure descriptor is analyzed as follows:

\textbf{Betweenness.} The time complexity of calculating a single-source shortest path in an unweighted graph via breadth-first search (BFS) \cite{bundy1984breadth} is $\mathcal{O}(m+n)$. It can be applied to compute the all-pairs shortest paths in a union subgraph with each node as the source, and the time complexity is $\mathcal{O}((m+n)n)$. For the whole graph, the complexity is $\mathcal{O}((m+n)nM)$.

\textbf{Count\_ne.} The complexity of counting the node number and edge number in a single union subgraph are $n$ and $m$, respectively. For the whole graph, the complexity is $\mathcal{O}((m+n)M)$.

\textbf{Curvature.} According to \cite{ye2019curvature}, the complexity of calculating the curvatures is $\mathcal{O}((d_u \cdot d_v)^{\omega})$, where $d_u$ and $d_v$ are the degrees of the nodes $u$ and $v$. Here we approximate both terms by $N$, then the complexity of calculating curvatures is $\mathcal{O}(N^{2\omega})$, in which $\omega$ is the exponent of the complexity of matrix multiplication (the best known is 2.373).


\textbf{Count\_cycle.} The complexity of counting $k$-tuple substructures, for example $k$-cycles, in the whole graph is $\mathcal{O}(N^k)$, where $k \geq 3$ in general.

\textbf{Ours.} The complexity of constructing a shortest path matrix and performing SVD for a single union subgraph are $\mathcal{O}((m+n)n)$ and $\mathcal{O}(n^3)$, respectively. So the total complexity is $\mathcal{O}(n^3 M)$.

Generally speaking, $M>N \gg m>n$. Thus, the complexity of our structure descriptor is slightly higher than that of Count\_ne and Betweenness, and significantly lower than that of Curvature and Count\_cycle.

\end{document}